\documentclass{article}

\usepackage{wrapfig} 
    \usepackage[preprint]{neurips_2025}

\usepackage[utf8]{inputenc} 
\usepackage[T1]{fontenc}    
\usepackage{hyperref}       
\usepackage{url}            
\usepackage{booktabs}       
\usepackage{amsfonts}       
\usepackage{nicefrac}       
\usepackage{microtype}      
\usepackage{xcolor}         
\usepackage[font=small]{caption} 
\usepackage{subfigure}

\usepackage{amsmath}
\usepackage{algorithmic}
\usepackage{algorithm}
\usepackage{array}
\usepackage{textcomp}
\usepackage{stfloats}
\usepackage{verbatim}
\usepackage{graphicx}
\usepackage{cite}
\usepackage{times}
\usepackage{epsfig}
\usepackage{amssymb}
\usepackage{multirow}
\usepackage{bbding}
\usepackage{makecell}
\usepackage[normalem]{ulem} 
\usepackage{balance}
\usepackage{caption,subcaption}
\usepackage{adjustbox}
\usepackage[capitalize]{cleveref}
\crefname{section}{Sec.}{Secs.}
\Crefname{section}{Section}{Sections}
\Crefname{table}{Table}{Tables}
\crefname{table}{Tab.}{Tabs.}
\usepackage{etoolbox}

\title{PSI: A Benchmark for Human Interpretation and Response in Traffic Interactions}

\author{
Taotao~Jing\textsuperscript{1} \qquad Tina~Chen\textsuperscript{2} \qquad Renran~Tian\textsuperscript{3}\thanks{Corresponding Author} \qquad Yaobin~Chen\textsuperscript{2} \qquad Joshua~Domeyer\textsuperscript{4} \\
\textbf{Heishiro~Toyoda\textsuperscript{4} \qquad Rini~Sherony\textsuperscript{4} \qquad Zhengming~Ding\textsuperscript{1}} \\
\textsuperscript{1}Tulane University \qquad \textsuperscript{2}Purdue University \\
\textsuperscript{3}North Carolina State University \qquad \textsuperscript{4}Toyota Motor North America \\
\{tjing, zding1\}@tulane.edu \qquad cchen.tina@gmail.com \quad rtian2@ncsu.edu \quad chen62@purdue.edu \\ 
\{joshua.domeyer, rini.sherony\}@toyota.com \quad heishiro.toyoda.tmc@tri.global
}

\begin{document}

\maketitle

\begin{abstract}
  Accurately modeling pedestrian intention and understanding driver decision-making processes are critical for the development of safe and socially aware autonomous driving systems. However, existing datasets primarily emphasize observable behavior, offering limited insight into the underlying causal reasoning that informs human interpretation and response during traffic interactions. To address this gap, we introduce PSI, a benchmark dataset that captures the dynamic evolution of pedestrian crossing intentions from the driver’s perspective, enriched with human-annotated textual explanations that reflect the reasoning behind intention estimation and driving decision making. These annotations offer a unique foundation for developing and benchmarking models that combine predictive performance with interpretable and human-aligned reasoning. PSI supports standardized tasks and evaluation protocols across multiple dimensions, including pedestrian intention prediction, driver decision modeling, reasoning generation, and trajectory forecasting and more. By enabling causal and interpretable evaluation, PSI advances research toward autonomous systems that can reason, act, and explain in alignment with human cognitive processes. The dataset is publicly available at \url{http://situated-intent.net/pedestrian\_dataset/}.
\end{abstract}

\section{Introduction}
\label{sec:intro}


As automation capabilities increase, autonomous vehicles (AVs) can perform their driving roles more easily in highway and freeway contexts. However, the complexity of mixed urban traffic, especially with pedestrians, remains challenging~\citep{dadvar2021california,boggs2020exploratory}. Interactions with pedestrians are safety-critical and affect traffic efficiency and user attitudes~\citep{domeyer2020vehicle}. To operate around pedestrians, most AVs make decisions in responding to their trajectories and optimize the ego motions to avoid crashes~\citep{li2021attentional, chen2021path}. Vehicle dynamics are often calculated based on the sensing inputs of motion parameters from the surrounding objects~\citep{wu2019dynamic}. 
To support vehicle decision-making during pedestrian encounters, learning-based approaches have been applied to predict pedestrian trajectories and actions~\citep{best2017autonovi,li2021attentional, herman2021pedestrian}, 
with an increasing number of algorithms and benchmark datasets published recently~\citep{chen2021survey, yang2022predicting, xu2022remember, guo2022end, girase2021loki}. However, pedestrian behavior prediction models have reduced performance for long duration predictions and still struggle to handle sudden action changes~\citep{bokkasam2025pedestrian, huang2024gpt, rasouli2024diving}. 

Driving is not only a technical skill but also a social skill~\citep{vinkhuyzen2016developing,pelikan2021autonomous}. This perspective emphasizes the importance of social factors like the goals and intentions of different road users and the role of personal experiences in understanding and planning behaviors during interactions. Similar to human drivers, AVs need to continuously coordinate with other human road users by exchanging intentions through communication~\citep{pelikan2021autonomous}.    
Thus, pedestrian intention prediction has been widely discussed in the literature, with many earlier studies using trajectories and actions (e.g., walking) as surrogates of actual pedestrian intention~\citep{girase2021loki,zhang2021pedestrian}. Considering that human intention reflects the degree of motivations to influence or act certain behaviors~\citep{ajzen1991theory} and there is an intention-behavior gap~\citep{sheeran2016intention}, i.e., people do not or cannot always do the things they intend to do, recent research recognizes the importance of directly studying pedestrian intentions in addition to the observable behaviors. It is expected that the prediction of intentions can advance the timing of accurately predicting actions, while also improving the accuracy of predicting long-term behavior and behavioral changing points. 

One pioneering study~\citep{rasouli2019pie} applied a crowd-sourcing method to label pedestrian crossing intentions. The authors asked data annotators to watch pre-recorded pedestrian videos and estimate the pedestrians' intentions to cross the street at one pre-determined critical frame for each video. Although the datasets support many models ~\citep{wang2022stepwise, yao2021bitrap}, there are several limitations that need to be addressed, including unclear intention definition, missing intention changes, label subjectivity, and lack of reasoning and background knowledge, which will be discussed in details in section 1.1.


\begin{figure*}[t]
\centering
\includegraphics[width=1\linewidth]{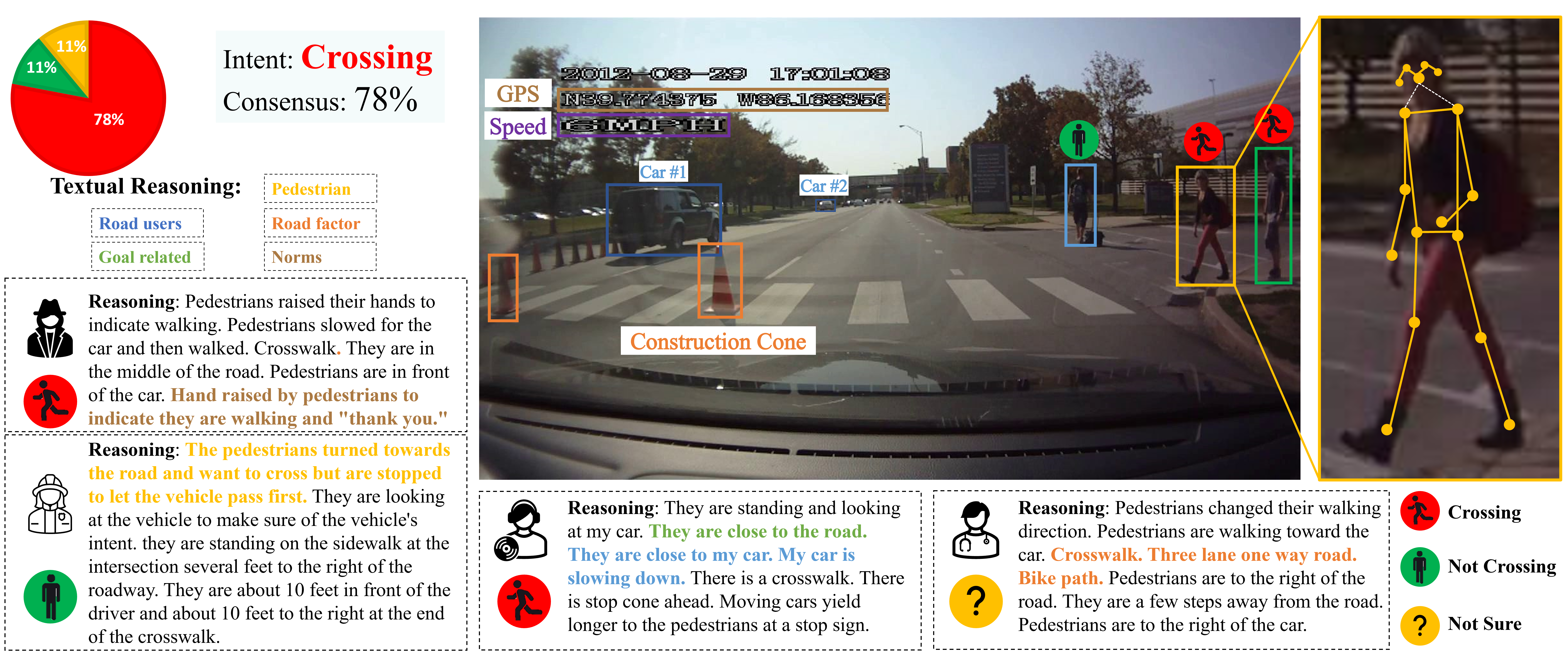}
\caption{Annotation examples: (1) bounding boxes and postures of pedestrians, crossing intent and textual reasoning from multiple annotators, road users, street structure (traffic lights, stop signs, construction cones, etc.), GPS, and speed; (2) textual reasoning of the intent estimation with different topics (pedestrians, road users, road factor, goal-related, social norms).}
\vspace{-8mm}
\label{fig:dataset_multimodal}
\end{figure*}

\subsection{Related Datasets}

In the last few years, there has been a sharp increase in the number of datasets released for pedestrian behaviors and automated vehicles research. These datasets are typically collected in major urban environments with pedestrian-centric annotations. 
Several datasets focus on or have labels related to pedestrian behaviors. The earlier data set, JAAD~\citep{kotseruba2016joint}, annotated pedestrian actions to study their crossing behavior with scene attributes also heavily annotated. STIP~\citep{liu2020spatiotemporal} contains pedestrian crossing action annotations, but lack additional attributes other than tracked bounding boxes. TITAN~\citep{tian2014estimation} classifies pedestrian actions into 43 sub-categories to try to explain their behavior. PedX~\citep{kim2019pedx} focuses on 3D human pose annotations to explain pedestrian behavior. Differently, Euro-PVI~\citep{bhattacharyya2021euro} is a dataset of pedestrian and bicyclist trajectories that encompasses diverse vehicle-pedestrian (and bicyclist) interactions. VIENA2~\citep{aliakbarian2018viena2} gathers a large-scale dataset covering various driving conditions and environments obtained using the GTA-V video game, and the data are annotated with a total of 25 distinct action classes for anticipatory driving action. 

For explainable AV datasets, normally there are both visual inputs and text annotations. Berkeley Deep Drive eXplanation (BDD-X) dataset~\citep{kim2018textual} is a subset of the BDD dataset~\citep{xu2017end} with action description and justification for all the driving events. BDD Object Induced Actions (BDD-OIA) dataset~\citep{xu2020explainable} is a subset of BDD100K~\citep{yu2018bdd100k}. 
To increase scene diversity, these videos were selected under various weather conditions and times of the day. This resulted in 22,924 five-second (5-s) video clips, which were annotated on MTurk for the 4 actions and 21 explanations. However, none of the existing datasets explains pedestrian crossing intents in naturalistic scenes.

Few earlier datasets address pedestrian intentions. The PIE~\citep{rasouli2019pie} dataset directly labels pedestrian crossing intentions using crowd-sourcing workers. Several other studies, like the LOKI~\citep{girase2021loki} dataset, use future pedestrian actions to surrogate current intentions. However, these datasets have some key limitations. 
\underline{First}, current annotations fail to distinguish between lower-level and higher-level intentions. Psychological literature defines two types of human intentions: future-directed intentions, which support planning and the formation of sub-intentions, and present-directed intentions, which directly lead to actions~\citep{bratman1984two, cohen1990intention}. In pedestrian behavior, future-directed intentions refer to the general plan to cross a road—useful for route planning but not necessarily indicative of immediate behavior—while present-directed intentions influence real-time actions such as stopping or crossing. Most existing datasets conflate these two, resulting in mixed or conflicting intention labels and predictions. \underline{Second,} while future-directed intentions typically remain stable during an encounter, present-directed intentions are highly dynamic and influenced by moment-to-moment contextual changes. Ethnographic studies have documented behavioral shifts during vehicle-pedestrian negotiations~\citep{dey2017pedestrian,sucha2017pedestrian}, but prior datasets only provide annotations at isolated critical frames, neglecting the evolving nature of real-time decision-making. This temporal gap introduces uncertainty in model training and evaluation when intention labels are assumed to hold across entire encounters.

\underline{More importantly}, crowd-sourced intention annotations often suffer from subjectivity. Disagreements among annotators are common in subjective labeling tasks~\citep{hube2019understanding}, and divergent estimations of pedestrian intentions have been noted in published datasets~\citep{rasouli2019pie}. Although soft labels~\citep{liu2019learning, wang2019classification} and annotation aggregation methods~\citep{tao2018domain, wallace2022debiased} can mitigate this issue, robust modeling of annotation uncertainty requires datasets with a large, demographically balanced pool of annotators—an element lacking in current datasets.
\underline{Finally}, existing datasets lack insight into the reasoning behind intention estimation. Pedestrian behavior is influenced by a rich array of contextual and behavioral cues~\citep{rasouli2017understanding, elahi2021novel}, yet most datasets exclude verbal reasoning. As foundation models and vision-language models (VLMs) have advanced capabilities in generalization, few-shot learning~\citep{brown2020language}, and in-context learning~\citep{dong2022survey}, incorporating human reasoning is critical to improve AI understanding and explainability~\citep{hou2020visual, kato2018compositional}. Verbal rationales from human drivers not only offer background knowledge but also serve as valuable resources for improving interpretability in autonomous systems~\citep{kim2018textual, zablocki2021explainability}.

\subsection{Objectives and Contributions}

This study introduces a new dataset to measure \textbf{Pedestrian Situated Intent (PSI)}, as illustrated in Figure~\ref{fig:dataset_multimodal}, defined as a pedestrian's intention to cross in front of the ego vehicle within a specific contextual interaction. PSI refers to the pedestrian's intention to cross the ego vehicle's path at a particular location before the vehicle arrives at that point. This definition focuses on immediate, present-directed crossing actions from the AV’s perspective, highlighting the associated risks and urgency. Unlike broader definitions of pedestrian intent as a general desire to cross the road, PSI provides more temporal and spatial boundaries, making it especially relevant for AV decision-making.

\begin{table*}[th!]
\vspace{-2mm}
\centering
\captionsetup{type=table}
\renewcommand{\arraystretch}{1.2}
\caption{Comparison of popular pedestrian behavior datasets in the AV research area. ${^{*}}$Number of annotated frames only include 2D annotations. 
${^{\dagger}}$Number of classes used for annotating objects in the dataset. ${^{\ddagger}}$Number of frames manually annotated. ${^{**}}$Dataset contains LiDAR but we do not consider it here. ``Disagr.'' denotes the disagreement across multiple observers for the same case.}\vspace{-2mm}
\label{tb:datasets}\small
\setlength{\tabcolsep}{0.5pt}
\begin{tabular}
{ccc|ccc|ccc|ccc}
\Xhline{1pt}
& & & \multicolumn{3}{c|}{Visual Annotations}  & \multicolumn{3}{c|}{Pedestrian Intent}  & \multicolumn{3}{c}{Driving Decision}  \\
\hline
\textbf{Dataset} & \textbf{Year~} & \textbf{\#Frames}$^{*}$  & \textbf{B-Boxes} & \textbf{Pose} & \textbf{Class}$^{\dagger}$ & \textbf{~Intent~}  & \textbf{Disagr.} & \textbf{~Reason~}  & \textbf{~Drive~} & \textbf{~Disagr.~} & \textbf{~Reason~}  \\
\Xhline{1pt}
Ours (PSI) & 2022 & 79k & \color{blue}\CheckmarkBold & \color{blue}\CheckmarkBold & 10 & 987k & \color{blue}\CheckmarkBold & \color{blue}\CheckmarkBold & \color{blue}\CheckmarkBold & \color{blue}\CheckmarkBold & \color{blue}\CheckmarkBold \\
\hline

LOKI~\citep{girase2021loki} & 2021 & 8k & \color{blue}\CheckmarkBold & \color{red}\XSolidBrush & 8 & 28k & \color{red}\XSolidBrush & \color{red}\XSolidBrush & \color{red}\XSolidBrush & \color{red}\XSolidBrush & \color{red}\XSolidBrush \\
\hline
STIP~\citep{liu2020spatiotemporal} & 2020 & 110k$^\ddagger$ & \color{blue}\CheckmarkBold & \color{red}\XSolidBrush & 1 & \color{red}\XSolidBrush & \color{red}\XSolidBrush & \color{red}\XSolidBrush & \color{red}\XSolidBrush & \color{red}\XSolidBrush & \color{red}\XSolidBrush \\
\hline
TITAN~\citep{malla2020titan} & 2020 & 75k & \color{blue}\CheckmarkBold & \color{red}\XSolidBrush & 3 & \color{red}\XSolidBrush & \color{red}\XSolidBrush & \color{red}\XSolidBrush & \color{red}\XSolidBrush & \color{red}\XSolidBrush & \color{red}\XSolidBrush \\
\hline
PIE~\citep{rasouli2019pie} & 2019 & 293k & \color{blue}\CheckmarkBold & \color{red}\XSolidBrush & 6 & 27k & \color{red}\XSolidBrush & \color{red}\XSolidBrush & \color{red}\XSolidBrush & \color{red}\XSolidBrush & \color{red}\XSolidBrush \\
\hline
PedX$^{**}$~\citep{kim2019pedx} & 2019 & 5k & \color{red}\XSolidBrush & \color{blue}\CheckmarkBold & 1 & \color{red}\XSolidBrush & \color{red}\XSolidBrush & \color{red}\XSolidBrush & \color{red}\XSolidBrush & \color{red}\XSolidBrush & \color{red}\XSolidBrush \\
\hline
JAAD~\citep{kotseruba2016joint} & 2016 & 82k & \color{blue}\CheckmarkBold & \color{red}\XSolidBrush & 3 & \color{red}\XSolidBrush & \color{red}\XSolidBrush & \color{red}\XSolidBrush & \color{red}\XSolidBrush & \color{red}\XSolidBrush & \color{red}\XSolidBrush \\
\hline

BDD-100K~\citep{yu2018bdd100k} & 2018 & 100k & \color{blue}\CheckmarkBold & \color{red}\XSolidBrush & 10 & \color{red}\XSolidBrush & \color{red}\XSolidBrush & \color{red}\XSolidBrush & \color{red}\XSolidBrush & \color{red}\XSolidBrush & \color{red}\XSolidBrush \\
\Xhline{1pt}
\end{tabular}
\vspace{-2mm}
\end{table*}

With this definition, PSI represents a temporally dynamic mental state that evolves with interaction context which is not captured in existing benchmark datasets. The new PSI dataset addresses key challenges in intention annotation, including subjectivity, human bias, reasoning background, and algorithm explainability, by increasing annotator diversity and collecting textual reasoning behind intention estimates. Intention annotations and reasoning were gathered from up to 24 annotators for each scene, who were drawn from varied age groups with balanced gender representation. The collected reasoning texts serve as a valuable knowledge base for AI decision-making. Additionally, the PSI dataset includes high-quality manual annotations of common visual features, including object detection and categorization, object tracking, and pedestrian postures. These rich annotations support a wide range of computer vision tasks using traditional learning techniques or most recent in-context learning, prompt engineering, and finetuning of foundation models.

We compare the new PSI dataset with several pedestrian-related dataset and show some key attributes as Table~\ref{tb:datasets}.
From the comparison, we observe that PSI dataset is the first dataset annotating the pedestrian crossing intention and driving decision together \textit{continuously} with reasoning descriptions and inter-driver disagreements (soft labels). It is noteworthy that the intention definition in the PSI dataset is present-directed which is different from all other crossing intention definitions~\citep{rasouli2019pie, kotseruba2016joint}. Because the PSI intention is annotated continuously across all the frames, the dataset has a much larger number of frames with intention annotations. These manually annotated frames provide more reliable intention labels for any sequences captured in the data, which are more accurate than the common practice of extending the intention annotation for one critical frame towards all the frames in the current pedestrian intention prediction research. To our best knowledge, disagreements among annotators and textual reasoning are explored for the first time in pedestrian scene understanding.



\begin{figure*}[t]
\centering
\includegraphics[width=1\linewidth]{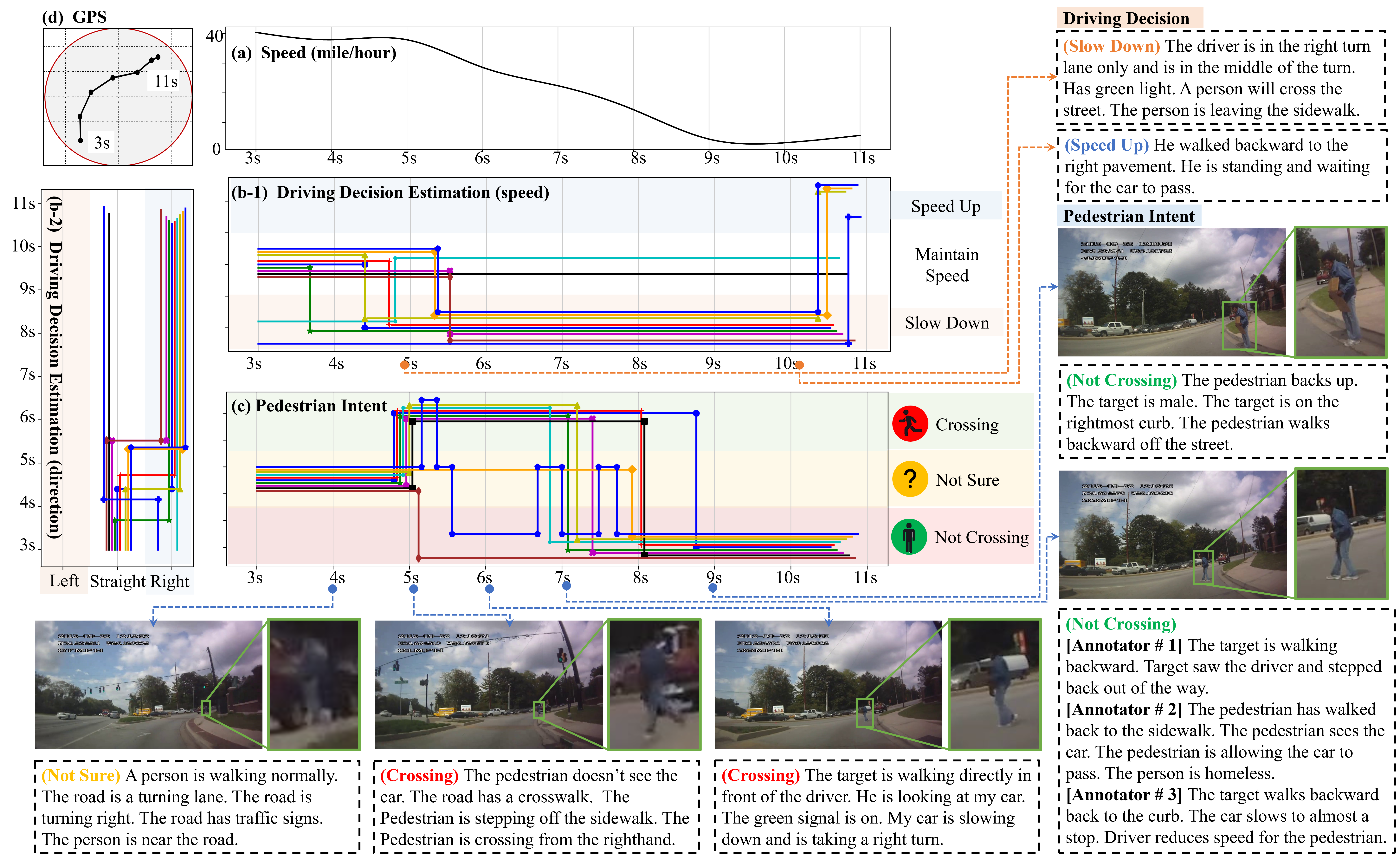}
  \caption{The example shows dynamic changes in estimated pedestrian situated intents and driving decisions (speed and direction), as well as reasoning descriptions from multiple annotators. Vehicle speed and GPS position are also included. The colored lines represent annotations from different annotators for the same case. Each annotator was asked to estimate the changes of the pedestrian’s crossing intent and justify their reasoning, as well as their driving decisions (direction and speed). This dynamic estimation process reflects the interaction between pedestrians and drivers, leading to consensus or disagreement among annotators. At times, all annotators agree on the same estimation. Conversely, there may be two or three opinions when there is disagreement. This diversity in the PSI dataset enhances the robustness of the annotations and provides a more comprehensive understanding of the scene. }
   \vspace{-5mm}
   \label{fig:dataset_intent}
\end{figure*}

\section{The PSI dataset}


\subsection{Data Preparation and Unique Features}

The PSI dataset contains 196 vehicle-pedestrian encounters with potential conflicts. Different from most pedestrian behavior benchmark datasets whose videos are continuously recorded in urban and downtown areas, the PSI dataset was randomly sampled from over 70,000 pedestrian encounters captured in the TASI 110-car naturalistic driving study~\citep{tian2014estimation}. In the driving study, 110 drivers were recruited to install a data collection system in each of their own cars. For one whole year, the positions of and the driving scenes in front of the subject's cars were continuously recorded. From 1.4 million miles of driving data, more than 70,000 pedestrian encounters were identified across all cars. 
Multiple data annotators manually checked all these pedestrian encounters to identify potential conflict cases. The potential conflict is defined as that at a particular timestamp during the encounter, crashes would happen if both the car and the pedestrian keep their instantaneous speed and directions~\citep{tian2014estimation}. A total of more than 3,000 potential conflict encounters were labeled in the whole data set, from which the 196 encounters were randomly selected for the PSI dataset. Each case is 15 seconds with scene videos at 30 fps, GPS coordinates at 1 fps, and vehicle speed at 1 fps.

Compared with other benchmark datasets, the PSI cases have the following unique features: \textbf{a}) The cases are more representative of the driving experiences of normal drivers in different road and environmental situations across the period of a whole year; \textbf{b}) Every case has potential conflicts from the human driver's perspective, indicating the existence of vehicle-pedestrian interactions and ensuring the focused pedestrians are relevant to the driving decision-making; \textbf{c}) All the cases contain longer and more complicated pedestrian behavior and action changes, making the prediction tasks more challenging and realistic; and \textbf{d}) The driving behaviors reflect more than 100 normal drivers to represent human driving decisions and corresponding pedestrian responses in the natural road environment. Because of all these features, the PSI dataset is a more challenging and realistic dataset to develop and test pedestrian intentions and behaviors to better support AVs. 


\subsection{Visual Annotations}


\begin{wraptable}[6]{r}{0.6\textwidth}  
\centering
\vspace{-5mm}
\caption{List of labels for visual annotations.}
\label{tb:labels}\vspace{-3mm}
\begin{center}
\begin{tabular}{|p{1.5cm}|p{6cm}|}
\hline
 \multicolumn{2}{|c|}{Bounding Box Labels} \\
 \hline
 Person & pedestrian, rider \\
 \hline
 Vehicle & car, bus, bicycle, semi-truck, motorcycle \\
 \hline
 Road & traffic sign, traffic light, construction cone \\
\hline
\end{tabular}\vspace{10 mm}
\end{center}
\end{wraptable}

There are two types of annotations in PSI: visual and cognitive annotations. 
Visual annotations include bounding boxes and pedestrian pose estimation. 
Bounding boxes and poses are all tracking enabled, so each object and pose is tracked for the entirety of the video clip. We annotated 10 classes of traffic objects and agents for bounding boxes (pedestrian, rider, car, bus, bicycle, semi-truck, motorcycle, traffic sign, traffic light, construction cone). Table~\ref{tb:labels} lists the classes annotated with bounding boxes. 
For the 196 video clips, there are 79,837 frames annotated with lower-level visual annotations. In total, there are $682,378$ bounding box annotations for traffic objects and agents, along with $71,259$ unique pose annotations for $391$ pedestrians, using the MS COCO format~\citep{lin2014microsoft}. The pose annotations comprise 17 different keypoints (classes) annotated with three values $(x, y, v)$, where $x$ and $y$ values denote the coordinates, and $v$ indicates the visibility of the key point (visible / not visible). There are $373$ pedestrians out of the $391$ were identified as key pedestrians that also have cognitive annotations.


\subsection{Cognitive Annotations}
The cognitive annotations include pedestrian situated intention (PSI), human reasoning descriptions, and driver behavior during the interaction. (\textbf{a}) \textit{Pedestrian Situated Intention} is segmented as polygonal chains (in Figure~\ref{fig:dataset_intent}) representing the estimated pedestrian intentions to cross in front of the ego-vehicle. The intentions can switch from three states, namely ``Cross'', ``Not Cross'', and ``Not Sure''. Each polygonal curve is the time-based estimation from one human driver/annotator, classifying the pedestrian's intention into one of the three categories in every frame. The vertices form the segmentation boundaries when pedestrian intention changes.  (\textbf{b}) \textit{Driver Behavior} is annotated in response to the dynamics of the pedestrian intentions, including speed and direction changes along time. Similar to the PSI labels, frame-by-frame driver behavior labels from one annotator can be shown as one polygonal curve, representing the estimated driving speed changes among ``Speed Up'', ``Maintain Speed'', and ``Slow Down'' or direction changes among `Left'', ``Straight'', and ``Right''. (\textbf{c}) \textit{Human Reasoning Descriptions} are collected as text explanations corresponding to all the vertices along all the pedestrian intention and driver behavior segmentation polygonal curves. These descriptions explain the most critical reasoning logic from human drivers/annotators during the preceding scene segments, and are collected via multiple prompts from five categories of pedestrian behavior influential factors~\citep{elahi2021novel} to stimulate in-deep thinking and explanations. 

\subsubsection{Annotator Statistics}
74 data annotators were recruited in this study. Each annotator labels at least 45 encountering scene videos through a video experiment to obtain these labels. All the annotators (44 males and 30 females) have valid US driving licenses and are from 19 to 77 years old. The diverse backgrounds of annotators can ensure the representativeness of pedestrian intent estimation results and reasoning descriptions. 

\subsubsection{Intention Annotation Process}\label{SectionAnnotation}



In order to obtain the cognitive annotations, a video experiment was conducted for 9 to 24
video annotators per video. 
To capture human cognition in estimating pedestrian intents and making driving decisions, we developed a browser-based data annotation platform that facilitated the annotation process. All 196 pedestrian encounter videos were presented to data annotators through the user interface. The videos would play and pause at the first frame when the pedestrian becomes visible, with the target pedestrian indicated by a red arrow. Annotators were tasked with estimating the pedestrian's intention to cross in front of the ego vehicle at that moment, based on preceding scenes, and providing a brief reasoning description to support their estimation. Several key functions include: 
\vspace{-18 pt}
\begin{itemize}
    \item A novel Point-and-Explain (PaE) UI that mimics how humans explain to one another. The interface allows annotators to refer to a visual objects and explain the related reasoning, and automatically links visual and verbal objects in the data. 
    \vspace{-3 pt}
    \item Multi-prompt reasoning prompts use open-ended and close-ended questions to stimulate the annotator to provide reasoning from different perspectives. 
\end{itemize}

After the first frame, annotators need to estimate the pedestrian intent frame by frame until the end of the video. If the pedestrian's intention changed at a particular frame, a segmentation boundary was inserted, and a reasoning description was required to explain the situation. Each time a change in intention estimation was identified, reasoning was described in five aspects: pedestrian-related factors, other road users, goal-related factors, road factors, and social norms. To guide annotators and encourage more detailed reasoning descriptions, separate questions were provided as instructions. 

Each video will be repeated twice to label pedestrian intents and driving decisions separately, following the same process. Details of the experiment platform and process can be found in~\citep{elahi2024mindread}.

\subsubsection{Cognitive Annotation Statistics}
In the PSI dataset, $74$ annotators have labeled the defined Pedestrian Situated Intents frame by frame to achieve a total of more than $987k$ 
intention estimations. There are $5,773$ 
Pedestrian Situated Intent segmentation boundaries in the entire data set across all the annotators and cases. On average, each case has $29.45$  
boundaries from all the annotators. Accordingly, there are $5,773$ segments in all the intention segmentation curves with the same number of reasoning descriptions. Each encounter has $9$ to $24$ 
polygonal segmentation curves from all the subjects. On average, each annotator has $1.2$ 
segments for each encounter case in estimating the situated intent. The average length of the reasoning is $251.7$  
characters. 


\subsubsection{Annotation Discrepancies}

Due to the involvement of multiple video annotators with diverse backgrounds, each encounter in the dataset exhibits both similarities and discrepancies in terms of intention estimation and reasoning descriptions. These similarities often reflect common sense and social norms, while the discrepancies primarily arise due to individual differences among the annotators. To quantify the level of agreement among the annotators or drivers in pedestrian intention estimation, we utilize the agreement score, which represents the highest percentage of annotators who agree on the same pedestrian intention at a particular timestamp. Conversely, the disagreement score is calculated as the complement of the agreement score (i.e., 1 minus the agreement score) for all frames and cases.

The analysis of the results reveals two significant findings. First, in over 58\% of the encountered scenes, the average disagreement score throughout the entire duration remains below 20\%. This indicates that in the majority of situations, most annotators are able to reach a consensus in estimating the defined PSI, reflecting a high degree of agreement. Second, more than 30\% of the cases exhibit a maximum disagreement score exceeding 50\%. This suggests that there are specific instances within many cases where more than half of the annotators fail to reach a consensus in their PSI estimations. These findings highlight the inherent complexity of real-world driving scenarios and emphasize the necessity of involving a large number of annotators to capture diverse perspectives.
To support algorithm development and evaluation, the PSI dataset includes the original (soft) labels provided by all the different annotators. This facilitates further research and allows researchers to focus on scenes with higher or lower agreement scores, which can serve as indicators of task difficulties.




\subsubsection{Intention Annotation Distribution}

\begin{figure}[t]
  \centering
  \subfigure[Number of intention annotations obtained for each target pedestrian in the PSI dataset.]
  {
    \includegraphics[width=0.46\linewidth]{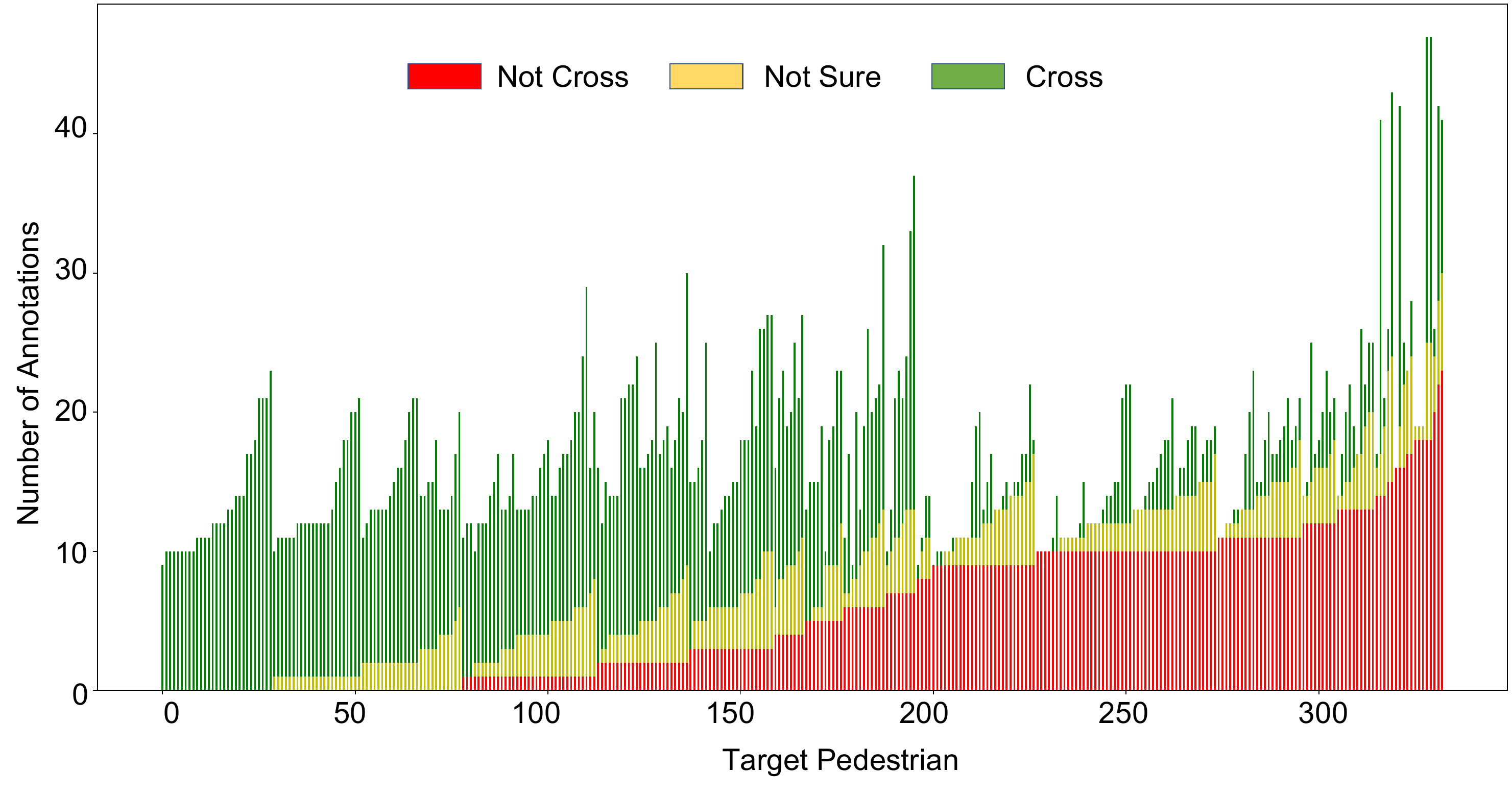}
    \label{fig:intent_count}
  }
  \hspace{0.03\linewidth}
  \subfigure[Time distribution (frame) of pedestrians annotated.]
  {
    \includegraphics[width=0.46\linewidth]{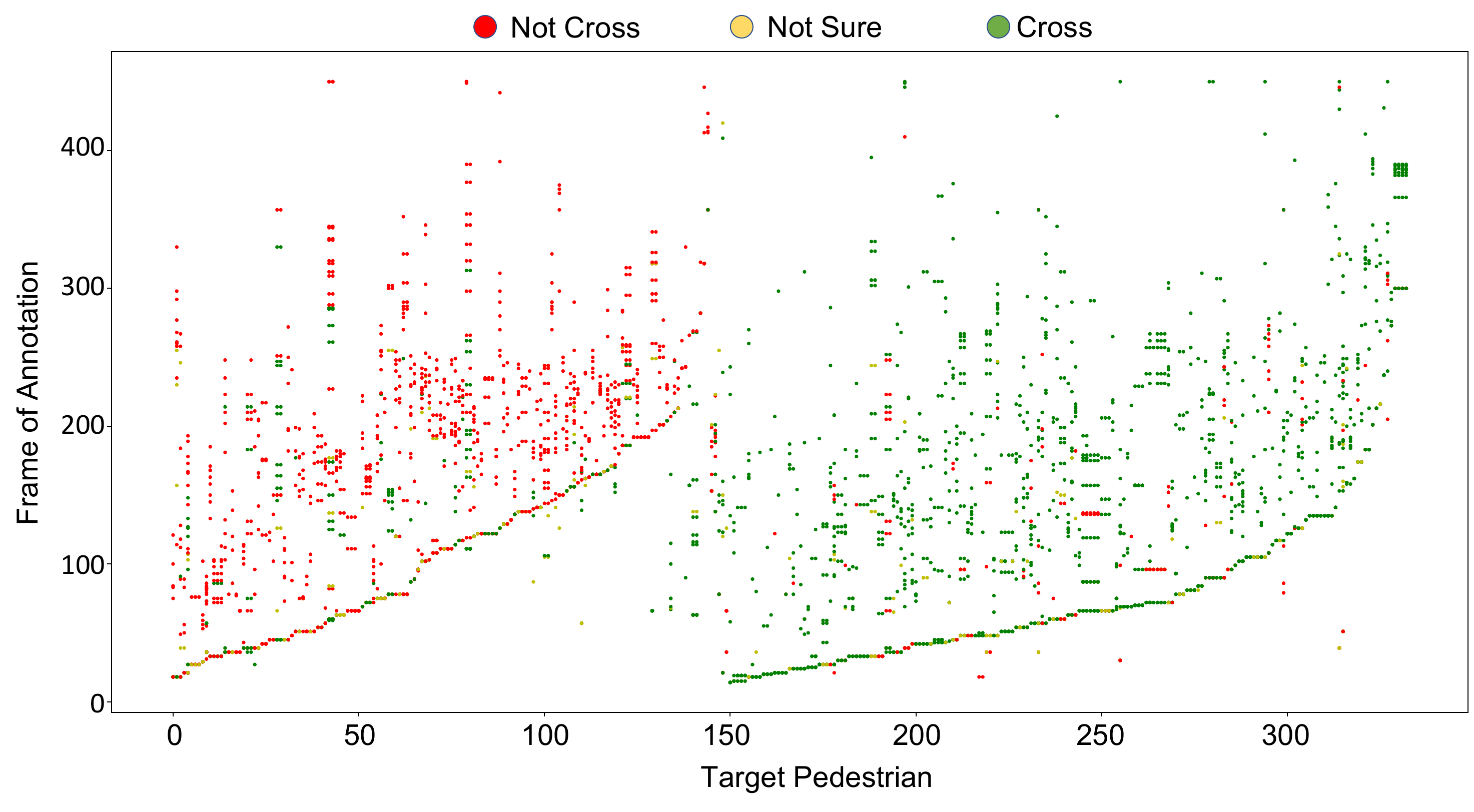}
    \label{fig:intent_distribution}
  }
  \caption{Overview of intention annotation statistics in the PSI dataset.}
  \label{fig:psi_annotation_stats}
  \vspace{-3mm}
\end{figure}

To explore the distribution of intention annotations in the PSI dataset, we present the number of intention annotations for each target pedestrian in Figure~\ref{fig:intent_count}. It should be noted that certain videos in the dataset contain multiple pedestrians, resulting in a total of over 300 target pedestrians. The distribution provides insights into the number of annotations for the "Cross" and ``Not Cross'' groups, as well as the level of disagreement among annotators concerning each target pedestrian.

Moreover, we observed that annotators tended to identify different key moments as triggers for changing their estimations of the crossing intention of target pedestrians. This implies that annotators believed the crossing intention of pedestrians in the present moment could change at various timings. Additionally, the crossing intention of certain pedestrians exhibited shifts during their interaction with vehicle drivers. To illustrate this variability, Figure~\ref{fig:intent_distribution} depicts the distribution of key moments (sequential frame numbers) for each target pedestrian, where intention annotations changed according to different annotators. The results demonstrate that annotators identified these key moments differently, and some pedestrians' crossing intentions indeed changed over the course of their interaction with vehicle drivers.


   

\section{Experimental Results}
\subsection{Reasoning with Large-scale Multimodal Models (LMM)}
\begin{figure}[t]
  \centering
\includegraphics[width=1\linewidth]{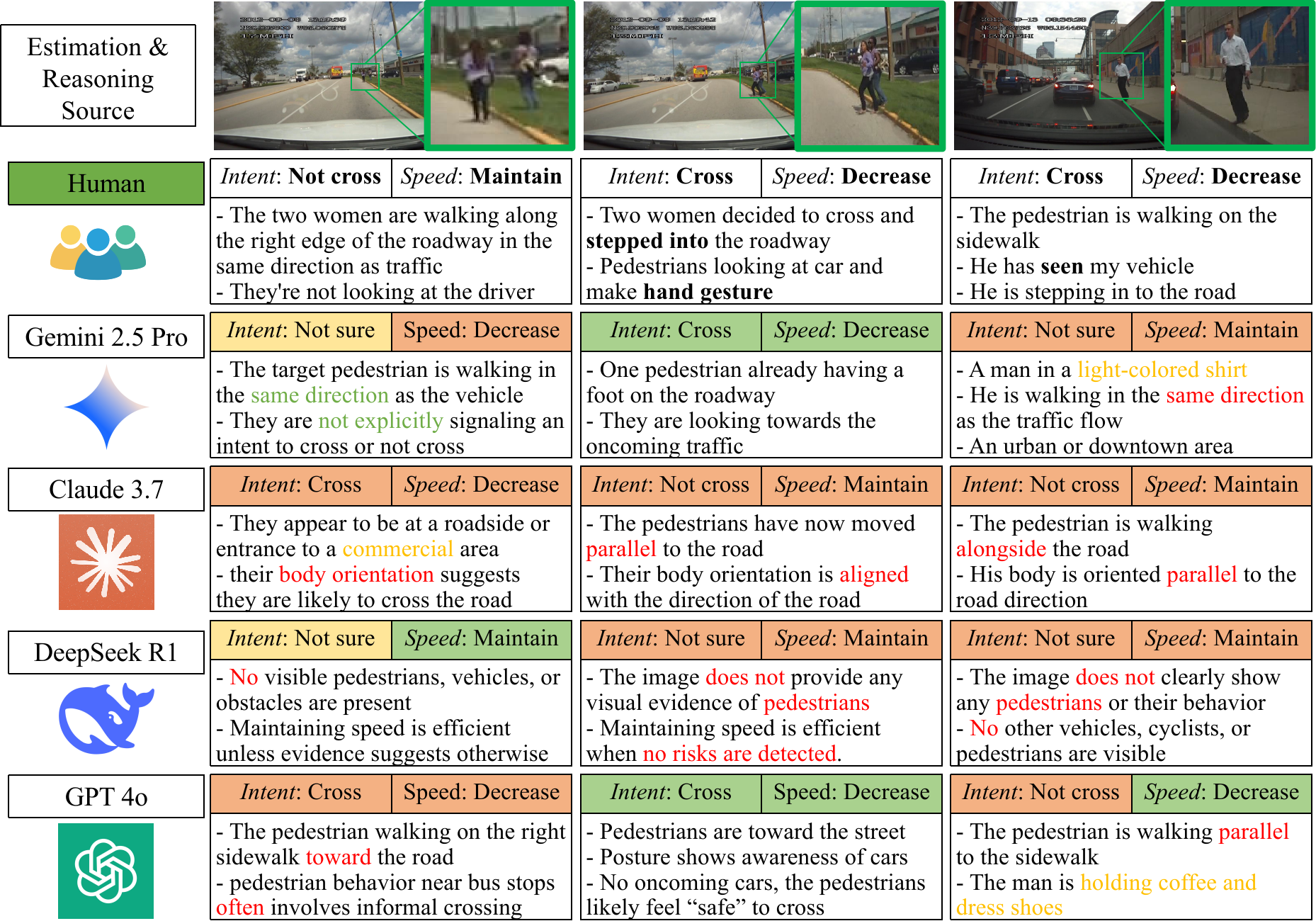}
  \caption{Comparison of pedestrian intent, driving decision estimations, and reasoning between human annotators and state-of-the-art AI models, all given the same prompts. For intent and driving decision: \textcolor{green}{green} indicates correct predictions matching human annotations, \textcolor{orange}{orange} denotes incorrect predictions, and \textcolor{yellow}{yellow} marks differing but reasonable predictions. For reasoning: \textcolor{green}{green} highlights relevant and accurate justifications, \textcolor{red}{red} indicates incorrect or flawed reasoning, and \textcolor{yellow}{yellow} shows irrelevant information.
  }
  \label{fig:reasoning}
\end{figure}

To highlight the importance of the reasoning annotations provided in the PSI dataset, we conducted a series of experiments on pedestrian intent prediction and driving decision estimation using several state-of-the-art large-scale multimodal AI models (LMM). Selected results are presented in Figure~\ref{fig:reasoning}. The AI models were prompted with the same instructions given to human annotators, guiding both estimation and reasoning processes. Both the models and human annotators were encouraged to justify their estimations from multiple perspectives, including the behavior of other road users, road layout and traffic signals, and human interaction norms. Since the evaluated models do not support video input, we provided a sequence of keyframes from each video and asked the models to perform estimation and reasoning based for the final frame.

\textbf{Pedestrian Intent and Driving Decision.} From the results, we observe that the performance of all evaluated models is inconsistent, particularly in complex urban scenarios or when the target pedestrian is positioned far from the ego-vehicle, as shown in the rightmost two columns of Figure~\ref{fig:reasoning}. Furthermore, we find that the models' estimations of pedestrian crossing intent and the corresponding driving decision are often not aligned. For example, in the first and third columns, Gemini 2.5 Pro~\citep{gemini25} expresses uncertainty about whether the pedestrian will cross in front of the ego-vehicle, yet it produces two different driving decisions in response-highlighting a lack of internal consistency in reasoning.

\begin{table*}[!t]
\centering
\caption{Comparison of Reasoning of Pedestrian Intent Prediction using LMM}
\setlength{\tabcolsep}{4pt} 
\scriptsize
\renewcommand{\arraystretch}{1.1}
\begin{tabular}{@{}l|ccccc@{}}
\hline
\textbf{Model} & \textbf{CIDEr ↑} & \textbf{BLEU-4 ↑} & \textbf{ROUGE ↑} & \textbf{METEOR ↑} & \textbf{BERTScore ↑} \\
\hline
Qwen-VL (7B)~\citep{bai2023qwen-vl} & 0.0216 & 0.0272 & 0.1786 & 0.1891 & 0.8557 \\
InternVL (8B)~\citep{chen2024internvl} & 0.0230 & 0.0337 & 0.1829 & 0.1806 & 0.8548 \\
SmoVLM (2.2B)~\citep{marafioti2025smolvlm} & 0.0227 & 0.0275 & 0.1797 & 0.1900 & 0.8558 \\
LLaMA 3.2 Vision (11B)~\citep{dubey2024llama} & 0.0194 & 0.0250 & 0.1935 & 0.1398 & 0.8625 \\
\hline
\end{tabular}
\label{table:reasoning}
\vspace{-2mm}
\end{table*}

\textbf{Reasoning Generation.}
Further challenges arise when the evaluated models attempt to generate reasoning to justify their estimations. One major issue is inaccurate perception of pedestrians and other small targets. All tested models struggled to determine the direction the target pedestrian was moving, which is a critical cue for intent prediction. DeepSeek R1~\citep{guo2025deepseek}, in particular, failed most tasks, often unable to detect any pedestrians in the scene, resulting in randomly generated estimations and explanations based purely on general knowledge rather than visual evidence. Additionally, the models frequently produced overly verbose contextual descriptions, often fixating on irrelevant visual details. For example, GPT-4o~\citep{chatgpt4o} remarked that a pedestrian was ``holding coffee,'' while Gemini 2.5 Pro noted the individual was ``in a light-colored shirt.'' Although factually correct, such observations are not informative for the task of pedestrian intent or driving decision estimation. In some cases, models also exhibited hallucinations, relying on pre-trained priors instead of the actual input images. For instance, in the first column, GPT-4o correctly noted that the pedestrian was walking ``toward the road'' but still predicted they will ``cross'', justifying this with the assumption that ``pedestrian behavior near bus stops often involves informal crossing'', which is an inference not clearly supported by the visual context.

Moreover, we conducted a reasoning generation task for pedestrian intent estimation using the PSI dataset. We evaluated several pre-trained vision-language models (VLMs), including InternVL, Qwen2.5-VL, and other baselines, on the test set. Each model received the same image and a standardized prompt generated by ChatGPT-4o, aiming to predict the pedestrian’s crossing intent and generate corresponding reasoning. We assessed performance using standard language generation metrics—CIDEr, BLEU-4, and ROUGE—providing a quantitative comparison across models and report the results in Table~\ref{table:reasoning}. These results establish baseline capabilities of current VLMs and highlight the challenges of generating accurate, context-aware reasoning for pedestrian intent.

These findings underscore the critical and complementary role of textual reasoning annotations provided by human annotators. Human-generated explanations are consistently more accurate, concise, and focused, avoiding distractions from irrelevant details and instead emphasizing subtle yet crucial visual cues and social interaction norms. Importantly, such reasoning often reflects human social understanding that goes beyond direct scene description. For example, in the second column, annotators inferred the pedestrian’s intent to cross based on subtle behaviors, such as making eye contact with the driver and gesturing, demonstrating the depth of natural reasoning. This highlights the central motivation of our work: leveraging human-like reasoning to complement visual input and improve the reliability and interpretability of pedestrian intent and driving decision estimation.

\begin{table*}[t]
\caption{Comparison of intention prediction results (\%)}
\vspace{-3mm}
\scriptsize
\begin{center}
\setlength{\tabcolsep}{6pt} 
{
\begin{tabular}
{l|cccc}
\Xhline{1pt}
{Baseline}  & Input & {Acc} & {F1} & {Acc$_\mathbf{avg}$} \\ 
\Xhline{1pt}
LSTM~\citep{hochreiter1997long} & box & 51.49$_{\pm 0.84}$ & 48.88$_{\pm 1.42}$ & 54.02$_{\pm 0.91}$ \\
Transformer~\citep{vaswani2017attention} & box & 56.66$_{\pm 0.44}$ & 50.77$_{\pm 1.18}$ & 53.91$_{\pm 0.39}$ \\
PIE~\citep{rasouli2019pie} & box+video & 59.04$_{\pm 0.14}$ & 56.13$_{\pm 0.09}$ & 61.93$_{\pm 0.08}$ \\
\midrule
LSTM+\textit{disagr.} & box & 52.75$_{\pm 0.12}$ & 49.60$_{\pm 0.09}$ & 56.74$_{\pm 0.15}$ \\
Transformer+\textit{disagr.} & box & 59.04$_{\pm 1.11}$ & 54.67$_{\pm 0.38}$ & 59.67$_{\pm 0.45}$ \\
PIE+\textit{disagr.} & box+video & 59.49$_{\pm 0.07}$ & 56.50$_{\pm 0.18}$ & 62.26$_{\pm 0.15}$\\
\midrule
\textbf{eP2P} & box+video & 60.30$_{\pm 0.18}$ & 57.01$_{\pm 0.08}$ & 62.26$_{\pm 0.11}$ \\
\textbf{eP2P}+\textit{disagr.} & box+video & 61.67$_{\pm 0.07}$ & 57.92$_{\pm 0.05}$ & 62.39$_{\pm 0.03}$ \\
\textbf{eP2P}+\textit{disagr.}+\textit{rsn} & box+video & \textbf{63.09}$_{\pm 0.16}$ & \textbf{58.55}$_{\pm 0.03}$ & \textbf{62.53}$_{\pm 0.03}$ \\
\bottomrule
\end{tabular}
}
\end{center}
\label{table:intent}
\vspace{-4mm}
\end{table*}

\subsection{Pedestrian Intent and Trajectory Prediction}
Given the proposed dataset, we will perform two tasks including pedestrian intent prediction and pedestrian trajectory prediction, to demonstrate the potential tasks that can be completed with the dataset and the use cases. 

\textbf{Explainable Pedestrian Trajectory Prediction (\textbf{eP2P}).} We developed an explainable Pedestrian Trajectory Prediction  model to complete multiple tasks, and the model consists of two modules, \textit{Pedestrian Situated Intent Prediction} and \textit{Trajectory Distribution Prediction}. The intent prediction module consists of a visual encoder and a text encoder to extract the corresponding features from the input video and textual reasoning annotations, as well as a global-local context feature fusion network to aggregate the global information of the whole scene and local cues from the region near the target pedestrian. Besides, an intent predictor and a reasoning generator are deployed to predict the crossing intent and the corresponding reason for the prediction, respectively. 
For the trajectory prediction module, an encoder and decoder are used to integrate the spatial-temporal information of the observed moving location of the pedestrian and the crossing intent estimation. Subsequently, the trajectory predictor predicts the future locations of the target pedestrian, while the uncertainty predictor estimates hyper-parameters for the model uncertainty.

\textbf{Experiments Setup and Metrics.} All 196 videos in the PSI dataset are split into three subsets, training/validation/test, by $0.75 : 0.05 : 0.2$. We sample clips of frame length $60$ with an overlap ratio of 0.9\citep{rasouli2019pie}. For intention prediction, with $0.5$ second (15 frames) observation as input, we predict the target pedestrian intention at the $16^{\text{th}}$-frame. For trajectory prediction, with the observed $0.5$ second sequence, we predict the target pedestrian's trajectory distributions for the following $0.5$, $1.0$, and $1.5$ seconds. For intent prediction, we report the overall accuracy (Acc), \textit{macro} average F1 score (F1), and class-wise average accuracy ({Acc$_\mathbf{avg}$}). In particular, the class-wise average accuracy is computed as the average accuracy across both ``corss'' and ``not cross'' cases. Contrasted with the overall accuracy, denoted as Acc, Acc$_\mathbf{avg}$ offers a more robust evaluation of model performance, particularly in scenarios where the test data exhibits imbalance.
For trajectory prediction, we evaluate the trajectory prediction using the metrics ADE/FDE and C$_\text{ADE}$/C$_\text{FDE}$.

More details of the proposed framework, experiments setup and implementation can be found in the \textit{supplementary} material.

\begin{table*}[t]
\caption{Comparison of errors (pixels) of different baselines on PSI datasets for pedestrian trajectory prediction}
\vspace{-3mm}
\scriptsize
\begin{center}
\linespread{1.1} 
\centering
\setlength{\tabcolsep}{2pt} 
\renewcommand{\arraystretch}{1} 
{
\begin{tabular}{ccccccccccccc}
\toprule
\multirow{2}{*}{Method} & \multicolumn{4}{c}{0.5s} & \multicolumn{4}{c}{1.0s} & \multicolumn{4}{c}{1.5s}  \\
\cmidrule(lr){2-5}
\cmidrule(lr){6-9}
\cmidrule(lr){10-13}
& $ADE$ \textcolor{green}{$\downarrow$} & $FDE$\textcolor{green}{$\downarrow$} & $C_{ADE}$\textcolor{green}{$\downarrow$} & $C_{FDE}$\textcolor{green}{$\downarrow$} & $ADE$ \textcolor{green}{$\downarrow$} & $FDE$\textcolor{green}{$\downarrow$} & $C_{ADE}$\textcolor{green}{$\downarrow$} & $C_{FDE}$\textcolor{green}{$\downarrow$}  & $ADE$ \textcolor{green}{$\downarrow$} & $FDE$\textcolor{green}{$\downarrow$} & $C_{ADE}$\textcolor{green}{$\downarrow$} & $C_{FDE}$\textcolor{green}{$\downarrow$}  \\
\midrule
LSTM~\citep{hochreiter1997long} & 31.65 & 54.75 & 21.54 & 37.71 & 57.80 & 110.90 & 39.82 & 76.96 & 86.98 & 176.74 & 60.18 & 122.69 \\
PIE~\citep{rasouli2019pie} & 24.26 & 33.97 & 15.55 & 22.31 & 35.80 & 62.35 & 23.53 & 41.86 & 52.61 & 109.45 & 35.15 & 74.54 \\
BiTraP-D~\citep{yao2021bitrap} & 25.34 & 31.57 & 16.67 & 20.83 & 35.14 & 60.37 & 23.32 & 40.71 & 52.13 & 111.58 & 35.11 & 76.49 \\
SGNet~\citep{wang2022stepwise} &24.08 & 40.22 & 15.32 & 26.74 & 43.72 & 85.18 & 29.03 & 57.81 & 67.38 & 143.01 & 45.41 & 97.67 \\
SGNet+\textit{evi}~\citep{wang2022stepwise} & 22.24 & 38.93 & 14.10 & 25.63 & 41.12 & 79.85 & 27.07 & 53.71 & 63.49 & 136.19 & 42.46 & 92.52 \\
BiTraP-NP~\citep{yao2021bitrap} & 19.11 & 27.83 & 12.38 & 18.37 & 31.91 & 60.18 & 21.17 & 40.71 & 48.37 & 102.54 & 32.52 & 69.74 \\
\midrule

\textbf{eP2P} & \textbf{18.91 } & \textbf{ 27.38 } & \textbf{ 10.87 } & \textbf{ 16.68 } & \textbf{ 30.02 } & \textbf{ 55.48 } & \textbf{ 18.50 } & \textbf{ 36.05 } & \textbf{ 45.55 } & \textbf{ 97.00 } & \textbf{ 29.24 } & \textbf{ 64.85} \\
\bottomrule
\end{tabular}
}
\end{center}
\label{table:trajectory}
\vspace{-4mm}
\end{table*}
\subsubsection{Results Comparison}

\textbf{Pedestrian Situated Intent Prediction.}
The results of pedestrian situated intent prediction are reported in Table~\ref{table:intent}. We compare our model with several baselines including LSTM~\citep{hochreiter1997long}, Transformer~\citep{vaswani2017attention}, and PIE~\citep{rasouli2019pie} with different inputs and annotations to calculate the learning objectives. Specifically, the baselines LSTM and Transformer only take the bounding boxes of the observed target pedestrian as input, while PIE leverages the local visual images in addition. The baselines are trained using the binary cross-entropy loss. In contrast, our proposed model incorporates both global and local observations as input knowledge and constrains the model training with both intent and reasoning annotations. We also compare the impact of utilizing the disagreement among all annotators to mitigate the effects of ambiguous situations.

Comparing PIE with the LSTM and Transformer baselines, we can observe how the visual observation of the local region, including the appearance of the target pedestrian, can enhance intent prediction accuracy significantly. When contrasting our \textbf{eP2P} model with PIE, we notice a performance improvement, F1 score increases from $56.13\%$ to $57.01\%$, due to the fusion of global-local contextual knowledge. Moreover, by incorporating the disagreement score among annotators to reweigh the learning objectives and mitigate distractions from uncertain cases, we observe further improvement in accuracy by $1.3\%$. Finally, integrating the reasoning generator module and training the model with textual annotations enhances the intent prediction accuracy by $1.4\%$, and the explanation generation module contributes to better human comprehension and verification.

\textbf{Trajectory Prediction.}
In Table~\ref{table:trajectory}, we compare our model with the baseline LSTM, which only inputs bounding boxes of the observed trajectory of the target pedestrian as input. Additionally, BiTraP estimates the long-term goal (end-point) of trajectories and introduces a novel bi-directional decoder to improve longer-term trajectory prediction accuracy, while SGNet estimates and uses goals at multiple temporal scales of trajectories. From the results, we observe that our model outperforms the compared baselines on all tasks, demonstrating the contribution of intent estimation and evidential uncertainty prediction to the trajectory prediction task. Moreover, we incorporate the evidential uncertainty prediction layer and corresponding training objectives into the SGNet framework and train the model with the same data, resulting in a boost in trajectory prediction accuracy (ADE@1.5s from 57.81 to 53.71). Such results further justify the effectiveness of the evidential uncertainty prediction design.

\section{Conclusion}
\label{sec:conclusion}
This paper introduced the Pedestrian Situated Intent (PSI) dataset, which captures pedestrian crossing intentions from an autonomous vehicle's perspective, with a focus on temporal and spatial context. A key innovation of PSI is its inclusion of human-annotated reasoning explanations, which provide deep insights into the cognitive processes behind pedestrian intentions and driver decision-making. These annotations go beyond mere behavior prediction, enabling models that not only forecast actions but also align with human reasoning. PSI supports critical tasks like intention prediction, trajectory forecasting, and reasoning generation, advancing the development of socially aware autonomous systems that can reason and act in a manner consistent with human thought processes. The dataset and its annotations are publicly available to drive further research and innovation in the field.

\newpage
\section{Acknowledgement}

This project was funded by the Toyota Collaborative Safety Research Center. This material is also based upon work supported by the National Science Foundation under Grant No.2145565.

\bibliographystyle{plain}
\bibliography{references}

\newpage
\section*{NeurIPS Paper Checklist}

\begin{enumerate}

\item {\bf Claims}
    \item[] Question: Do the main claims made in the abstract and introduction accurately reflect the paper's contributions and scope?
    \item[] Answer: \answerYes{} 
    \item[] Justification: The abstract and introduction clearly state the claims made, including the contributions to the PSI dataset and experiments in the paper.
    

\item {\bf Limitations}
    \item[] Question: Does the paper discuss the limitations of the work performed by the authors?
    \item[] Answer: \answerYes{} 
    \item[] Justification: We discussed the challenges of the data collection stage and the limitations of the dataset, as well as the limitations caused from existing AI models in our experiments in the manuscript.

\item {\bf Theory assumptions and proofs}
    \item[] Question: For each theoretical result, does the paper provide the full set of assumptions and a complete (and correct) proof?
    \item[] Answer: \answerNA{} 
    \item[] Justification: This is a dataset/benchmark work which does not have theory assumptions and proofs. Important references and motivations are properly cited.

    \item {\bf Experimental result reproducibility}
    \item[] Question: Does the paper fully disclose all the information needed to reproduce the main experimental results of the paper to the extent that it affects the main claims and/or conclusions of the paper (regardless of whether the code and data are provided or not)?
    \item[] Answer: \answerYes{} 
    \item[] Justification: All the dataset and processing code, experimental code and instructions are provided in the manuscript or open-source.

\item {\bf Open access to data and code}
    \item[] Question: Does the paper provide open access to the data and code, with sufficient instructions to faithfully reproduce the main experimental results, as described in supplemental material?
    \item[] Answer: \answerYes{} 
    \item[] Justification: All the dataset is host on HuggingFace, and corresponding codes are open-source on Github. 

\item {\bf Experimental setting/details}
    \item[] Question: Does the paper specify all the training and test details (e.g., data splits, hyperparameters, how they were chosen, type of optimizer, etc.) necessary to understand the results?
    \item[] Answer: \answerYes{} 
    \item[] Justification: In our paper, we detailedly described how the experimental results are obtained, and more codes including training and test process are provided on Github.

\item {\bf Experiment statistical significance}
    \item[] Question: Does the paper report error bars suitably and correctly defined or other appropriate information about the statistical significance of the experiments?
    \item[] Answer: \answerNA{} 
    \item[] Justification: We demonstrate the contribution of the new PSI dataset in this dataset/benchmark track paper, thus no statistical significance analysis of the experimental results are conducted.

\item {\bf Experiments compute resources}
    \item[] Question: For each experiment, does the paper provide sufficient information on the computer resources (type of compute workers, memory, time of execution) needed to reproduce the experiments?
    \item[] Answer: \answerYes{} 
    \item[] Justification: The experiments involved in this paper are based on public AI models such as GPT 4o or DeepSeek R1. Thus people can reproduce the results using those tools.
    
\item {\bf Code of ethics}
    \item[] Question: Does the research conducted in the paper conform, in every respect, with the NeurIPS Code of Ethics \url{https://neurips.cc/public/EthicsGuidelines}?
    \item[] Answer: \answerYes{} 
    \item[] Justification: We confirm that we follow the NeurIPS Code of Ethics in the whole process of this research paper.

\item {\bf Broader impacts}
    \item[] Question: Does the paper discuss both potential positive societal impacts and negative societal impacts of the work performed?
    \item[] Answer: \answerYes{} 
    \item[] Justification: Both positive and potential negative societal impact caused by the limitations of the PSI dataset is discussed.
    
\item {\bf Safeguards}
    \item[] Question: Does the paper describe safeguards that have been put in place for responsible release of data or models that have a high risk for misuse (e.g., pretrained language models, image generators, or scraped datasets)?
    \item[] Answer: \answerYes{} 
    \item[] Justification: All data collected in this dataset are thoroughly reviewed by legal teams and obtained necessary consent.

\item {\bf Licenses for existing assets}
    \item[] Question: Are the creators or original owners of assets (e.g., code, data, models), used in the paper, properly credited and are the license and terms of use explicitly mentioned and properly respected?
    \item[] Answer: \answerYes{} 
    \item[] Justification: All referenced or reused tools and codes are properly cited.

\item {\bf New assets}
    \item[] Question: Are new assets introduced in the paper well documented and is the documentation provided alongside the assets?
    \item[] Answer: \answerYes{} 
    \item[] Justification: Detailed documents and codes are provided alongside the dataset.

\item {\bf Crowdsourcing and research with human subjects}
    \item[] Question: For crowdsourcing experiments and research with human subjects, does the paper include the full text of instructions given to participants and screenshots, if applicable, as well as details about compensation (if any)? 
    \item[] Answer: \answerYes{} 
    \item[] Justification: Detailed instructions and data collection details are provided in the paper and open source assets.

\item {\bf Institutional review board (IRB) approvals or equivalent for research with human subjects}
    \item[] Question: Does the paper describe potential risks incurred by study participants, whether such risks were disclosed to the subjects, and whether Institutional Review Board (IRB) approvals (or an equivalent approval/review based on the requirements of your country or institution) were obtained?
    \item[] Answer: \answerYes{} 
    \item[] Justification: This work is conducted with Institutional Review Board approvals.

\item {\bf Declaration of LLM usage}
    \item[] Question: Does the paper describe the usage of LLMs if it is an important, original, or non-standard component of the core methods in this research? Note that if the LLM is used only for writing, editing, or formatting purposes and does not impact the core methodology, scientific rigorousness, or originality of the research, declaration is not required.
    \item[] Answer: \answerYes{} 
    \item[] Justification: The usage of LLM for writing polish and experiments are clearly mentioned.

\end{enumerate}

\end{document}